\documentclass{ecai}

\usepackage{latexsym}
\usepackage{amsmath,amssymb,amsfonts}
\usepackage{amsmath}
\usepackage{amsthm}
\usepackage{booktabs}
\usepackage{enumitem}
\usepackage{graphicx}
\usepackage{color}
\usepackage{bm}
\usepackage{relsize}
\usepackage{multirow}
\usepackage{import}
\usepackage{wrapfig}

\usepackage[dvipsnames]{xcolor}
\usepackage{xkcdcolors}

\usepackage[footnotesize]{caption}
\newcommand{\topcaption}[1]{\vspace{2mm}\caption{#1}\vspace{2mm}}
\newcommand{\bottomcaption}[1]{\vspace{2mm}\caption{#1}\vspace{6mm}}

\newcommand{\BibTeX}{B\kern-.05em{\sc i\kern-.025em b}\kern-.08em\TeX}

\begin{document}

\begin{frontmatter}

    \paperid{3954}

    \title{Calibration improves detection of mislabeled examples}

    \author[A]{\fnms{Ilies}~\snm{Chibane}}
    \author[A]{\fnms{Thomas}~\snm{George}}
    \author[A]{\fnms{Pierre}~\snm{Nodet}\thanks{Corresponding Author. Email: pierre.nodet@orange.com}}%
    \author[A]{\fnms{Vincent}~\snm{Lemaire}}

    \address[A]{Orange Research}

    \begin{abstract}Mislabeled data is a pervasive issue that undermines the performance of machine learning systems in real-world applications. An effective approach to mitigate this problem is to detect mislabeled instances and subject them to special treatment, such as filtering or relabeling. Automatic mislabeling detection methods typically rely on training a base machine learning model and then probing it for each instance to obtain a trust score that each provided label is genuine or incorrect. The properties of this base model are thus of paramount importance. In this paper, we investigate the impact of calibrating this model. Our empirical results show that using calibration methods improves the accuracy and robustness of mislabeled instance detection, providing a practical and effective solution for industrial applications.\end{abstract}

\end{frontmatter}

\section{Introduction}

Machine learning models are only as good as the data on which they are trained. Collecting and annotating large datasets requires substantial effort, and a prominent remaining issue is the quality of the labels: even popular research datasets such as MNIST or CIFAR-10/CIFAR-100 contain labeling issues \citep{northcutt2021pervasive}. Rather than going through the tedious process of carefully reviewing each label, some mislabeled detection techniques aim to automatically identify these instances \citep{frenay2013classification,guan2013survey}.

For each instance in the training set, these methods provide a scalar \emph{trust score}, indicating whether the provided label can be considered clean or noisy. Among these techniques, \emph{model-probing} detectors \citep{george2024mislabeled} exploit the fact that machine learning algorithms usually treat genuine examples differently from mislabeled ones. The difference in treatment is measured using \emph{probes}, a generic term that encompasses, for instance, the straightforward method of checking if the model's prediction aligns with the provided label \citep{wilson2007asymptotic}.

A remaining issue with such methods is to distinguish between genuine challenging examples and mislabeled ones. Such challenging examples are, for instance, encountered in underrepresented areas of the instance space; therefore, discarding them would imply that there are no more training examples to learn a correct decision boundary in this particular area. This issue is, for instance, found when dealing with class-imbalanced datasets or in tasks involving fairness issues where a prominent pattern is learned at the expense of correct prediction on a protected group.

\begin{figure}[t]
    \centering
    \includegraphics[width=0.6\linewidth]{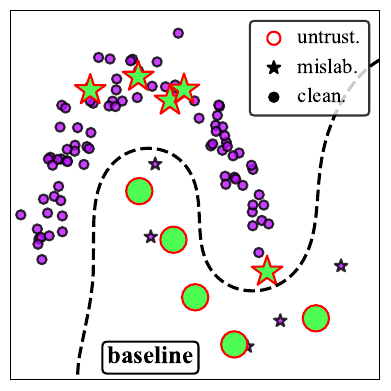}\\
    \includegraphics[width=0.6\linewidth]{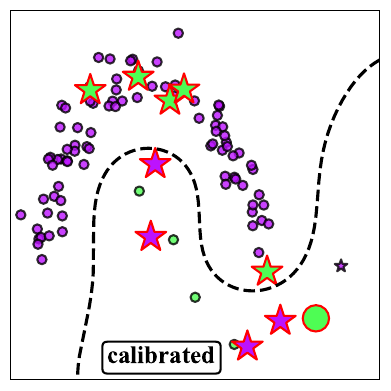}
    \bottomcaption{The 2-moons dataset with class imbalance, the lower moon \textcolor{xkcdLightNeonGreen}{$\bullet$} represents 10\% of the dataset and the top moon \textcolor{xkcdNeonPurple}{$\bullet$} 90\%. Five examples are mislabeled $\star$ for each of the two classes. The 10 examples with the lowest trust scores, obtained using (top) an uncalibrated (\textbf{baseline}) detector or (bottom) a \textbf{calibrated} detector are highlighted in \textcolor{red}{red}. For the uncalibrated detector, all examples from the minority class are untrusted; in contrast, the calibrated detector correctly flags examples from both classes as untrusted.}
    \label{fig:two_moons_pie}
\end{figure}

This comes from the fact that trust scores are built using base machine learning models, which are less accurate in more challenging areas. In general, they are, in particular, not well \emph{calibrated}: the confidence provided by the predictor does not reflect its expected accuracy on a particular instance. For detectors that use \emph{confidence}-based probes, where trust scores are derived from the base model’s predicted confidence, uncalibrated confidence may unevenly affect certain minority groups of examples, which might be flagged as mislabeled for the wrong reasons.

As a motivating example, in imbalanced classification datasets, trained machine learning models often exhibit under-confidence for minority class examples, leading to a higher likelihood of being considered mislabeled by detectors compared to majority class examples, for which models tend to be overconfident. Figure \ref{fig:two_moons_pie} illustrates this issue with a toy example (2 moons), and a confidence-based detector that relies on the predicted confidence of a model trained from the noisy and imbalanced dataset (see Appendix \ref{sec:motivating_examples} for details). The detector using non-calibrated (\textbf{baseline}) confidences assigns the lowest trust scores to the examples of the minority class, unlike the detector using \textbf{calibrated} confidences.

\paragraph{Related works} Alternatively to calibration, \citet{kuan2022model} propose to \emph{adjust} the predicted confidences to ensure that probes computed from these adjusted confidences evenly prioritize minority and majority classes. This adjustment is made by centering the distribution of confidences for each class to have a zero mean, and then transforming the centered confidences to a valid probability distribution summing to $1$. Although this method has shown empirical improvements, it remains heuristic.

To the best of our knowledge, the relationship between calibration and detection of mislabeled examples remains underexplored. The only (unpublished) work that we came across \citep{joel2023effect} studies the effect of label smoothing to i) improve prediction calibration and ii) mitigate the impact of noisy labels on model-probing detectors. However, it is not clear whether the claimed improvement comes from calibration, or if both are just desirable side effects of label smoothing. Furthermore, the study %
is limited to one detector \citep{pleiss_identifying_2020} and two datasets (CIFAR-10 and CIFAR-100 \citep{krizhevsky2009learning}).

\paragraph{Contribution} Our proposal in this paper is therefore to study the impact of \emph{calibrating} the base machine learning model of model-probing detectors. In particular:

\begin{itemize}
    \item We demonstrate that calibration prevents these detectors from wrongly assuming that examples from minority classes are mislabeled (Figure \ref{fig:minority_removed}).
    \item We show that detectors with calibrated base models significantly outperform uncalibrated detectors on a benchmark using a series of realistic noisy datasets with multiple model-probing detectors (Figure \ref{fig:plugins_clean_valid} and Table \ref{tab:results-logl}).
    \item We evidence the robustness of calibrating detectors under two practical constraints: noisy annotations and limited labeled examples (Figure \ref{fig:clean_vs_noisy} and \ref{fig:size_calib}).
\end{itemize}

\section{Background}
\label{sec:background}

In supervised classification, we want to predict a class for points sampled from a distribution $\mathbb{P}(x)$ on an input space $\mathcal{X}$ (e.g., the columns in a tabular dataset, or the pixels of an image). We restrict our study to the case where each point $x$ in the input space is associated with a single ground truth class $y_\text{true}(x)$ deterministically among $C$ possible ones. We have access to a training dataset $\mathcal{D}$ of samples from $\mathbb{P}$ and their corresponding labels $(x_i,y_i)$ where $y_i$ is the observed class that can be different from the ground truth class $y_\text{true}(x_i)$ when the labeling process is imperfect. The goal is to learn a classification model $\hat{f}(x)$ that takes an input example $x$ and returns a $C$-vector of confidences, where, $\forall c \in  [\![1, C]\!]$, $f(x)_c$ corresponds to the model's confidence that the instance belongs to the $c$ class. The predicted class $\hat{y}$ is determined by the class with the highest confidence: $\hat{y}(x) := \underset{c}{\arg\max} \, \hat{f}(x)_{c}$. The goal is to learn a model that generalizes well to unseen examples.

\subsection{Learning in presence of label noise}\label{sec:mislabeled}

The effectiveness of learning suitable functions $\hat{f}$ in supervised classification depends, among other factors, on the size of the training set. However, for each of these examples, it is necessary to annotate them, which can be a lengthy and costly process. This is why it is common to seek to automate this process.%

In the automated annotation toolbox lies: \emph{crowd labeling} \citep{yuen2011survey} which outsources the annotation to decentralized human annotators, \emph{web scraping} \citep{maas2011learning,xiao2015learning} which leverages query engine to quickly assemble datasets from online sources, or \emph{weak supervision} \citep{ratner2016data} which distills expert knowledge into hand-engineered labeling rules.

The drawback of all these tools is that the generated label $y_i$ might be different than the true label $y_{\text{true}}(x_i)$. For the case of crowd labeling, these errors can span from random human errors to examples genuinely hard to classify for humans.

In the most general case, the label noise is called noisy not at random (NNAR) as the probability of being mislabeled depends on the features and the label of the instance. When being mislabeled depends only on the label of the instance, the label noise is called at random (NAR) and when it does not depend on the label nor the features, the label noise is completely at random (NCAR) \citep{frenay2013classification}.

The \emph{noise transition matrix} $\mathbf{T}$ is the formal notation to distinguish between these forms of label noise. The noise transition matrix represents the probabilities from an instance $x_i$ to be assigned to class $y_i=\tilde{c}$ given belonging to the true class $y_{\text{true}}(x_i)=c$:
\begin{equation}
    \mathbf{T}_{\tilde{c}c}(x_i)=\mathbb{P}(y_i=\tilde{c}|y_{\text{true}}(x_i)=c)
\end{equation}

Getting back to our example with crowd labeling, random human errors fit in the NCAR category, and are probably easy to fix, meanwhile examples hard to classify for humans fit in the NNAR category, which seems harder to fix.

To address this drawback of automated annotation tools, a possible approach is to automatically detect the generated labeling errors in the training set \citep{frenay2013classification}.

\begin{figure*}[t]
    \centering
    \includegraphics[width=\linewidth]{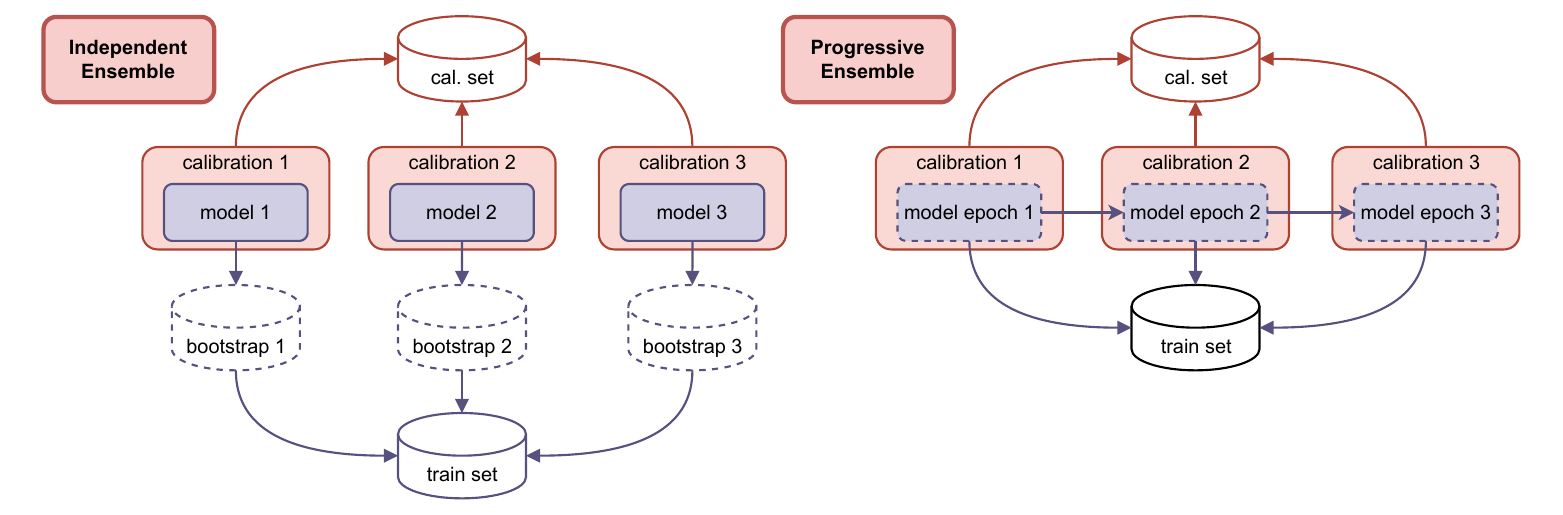}
    \bottomcaption{Schematic summary of calibration of model-probing detection methods with ensembling strategies. All models generated (or trained) by an ensembling strategy are independently calibrated on the same held-out calibration (cal.) set before probing. }
    \label{fig:model_probing_calibration}
\end{figure*}

\subsection{Model-probing methods for detection of mislabeled instances}\label{sec:model_probing}

Automated mislabeled detection methods give a \emph{trust score} to every pair instance/label in the training set. Less trusted examples can then be more carefully reviewed by active relabeling, or fully discarded as they are likely mislabeled (filtering). We focus on the family of \emph{model-probing} detection methods, since they encompass most state-of-the-art methods. We shortly describe their inner working next, and we refer the interested reader to \citep{george2024mislabeled} for a recent, in-depth discussion of model-probing methods.

These rely on fitting a machine learning model using training set examples, then \emph{probing} this model to obtain a hint on how regular or irregular a given label is for its corresponding instance, by means of a scalar trust score. The rationale is that robust machine learning models should behave differently for learning the signal carried by correctly labeled examples and the noise in the form of the subset of mislabeled examples. Probing is achieved in a variety of ways, including simple ideas (such as the predicted class by the model, its probability using a softmax, or the corresponding loss on a training example) or more intricate techniques (such as using the gradient with respect to the input of the model). Moreover, these detectors can be improved by using \emph{ensembling}, by probing the model at checkpoints during training, such as in boosting methods and in iteratively trained deep neural networks (progressive ensembles), or by training multiple models from different bootstrapped samples of the training set (named independent ensembles). If ensembling is used, the trust scores computed on all sub-models must be \emph{aggregated} to obtain a scalar trust score. All model-probing detectors can be described by these three building blocks: probe, ensemble, and aggregation.

A popular example that fits into this framework is the Area Under the Margin (AUM) method \citep{pleiss_identifying_2020}, where a deep learning predictor $\hat{f}$ is probed using individual confidence margins, computed on every training instance $(x_i, y_i)$ at several checkpoints $t$ during training (composed of $T$ checkpoints), then these margins are averaged to obtain a single trust score for every training example:
\begin{equation}
    \text{AUM}(x_i, y_i) = \frac{1}{T}\sum_{t=1}^T\left(\hat{f}_t(x_i)_{y_i} - \underset{c\neq y_i}{\max} \, \hat{f}_t(x_i)_{c}\right)
\end{equation}
Large margins during most of the training procedure indicate trusted examples, whereas low or negative aggregated margins are a sign of low confidence or incorrect labels.

In Section \ref{sec:benchmark}, we also experiment with other model-probing detectors, such as CleanLab \citep{northcutt2021confident} and Consensus \citep{jiang_characterizing_2021}, which rely on out-of-bag predicted confidences of an ensemble of $B$ models $\hat{f}_b$ trained on subsets $\mathcal{B}_b$ of the whole training set:
\begin{align}
    \text{CleanLab}(x_i, y_i)  & = \frac{1}{B}\sum_{\substack{b=1 \\(x_i, y_i)\notin \mathcal{B}_b}}^B\hat{f}_b(x_i)_{y_i}\\
    \text{Consensus}(x_i, y_i) & = \frac{1}{B}\sum_{\substack{b=1 \\(x_i, y_i)\notin \mathcal{B}_b}}^B \bm{1}_{\underset{c}{\arg\max} \, \hat{f}_b(x_i)_{c} = y_i}
\end{align}
and Small Loss \citep{jiang2018mentornet_smallloss}, which computes the per-sample loss, here the log-loss for classification, on the model $\hat{f}$ trained from the whole training set:
\begin{equation}
    \text{Small Loss}(x_i, y_i) = -\log\left(\hat{f}(x_i)_{y_i}\right)
\end{equation}

The three corresponding building blocks of the latter detectors are summarized in the following Table \ref{table:detectors}:
\begin{table}[h]
    \topcaption{Summary of benchmarked model-probing detectors.}
    \centering
    \begin{tabular}{llll}
        \toprule
        \textbf{Detector [Ref]}                         & \textbf{Probe} & \textbf{Ensemble} & \textbf{Aggregation} \\
        \midrule
        AUM \citep{pleiss_identifying_2020}             & margin         & progressive       & sum                  \\
        CleanLab \citep{northcutt2021confident}         & confidence     & independent       & mean (out-of-bag)    \\
        Consensus \citep{jiang_characterizing_2021}     & accuracy       & independent       & mean (out-of-bag)    \\
        Small Loss \citep{jiang2018mentornet_smallloss} & loss           & no ensemble       & not applicable       \\
        \bottomrule
    \end{tabular}
    \label{table:detectors}
\end{table}

The performance at detecting mislabeled examples crucially depends on the choice of the machine learning model, which is at the heart of the detection method, including its robustness to noisy examples. \emph{Calibration} of this model, viewed as a way to obtain more reliable predictions, is therefore a promising candidate for improved detection of mislabeled examples.

\subsection{Calibration of machine learning models}
\label{sec:calibration}

In classification, it is common practice to only use the predicted class returned by a machine learning model to make a decision. However, in some cases, we are also interested in the confidence given to each class for a given instance. But by default, off-the-shelf classifiers provide unreliable confidences because there is no guarantee that they are \emph{calibrated}.

Informally, a machine learning model is said to be calibrated if its confidence for each class corresponds to the observed frequency for this class to be correct \citep{zadrozny2002transforming}. Formally, a classifier is classwise calibrated \citep{perez2023beyond} if for all confidence thresholds $\tau$ in $[0,1]$ and for all classes $c$ in $[\![1, C]\!]$:

\begin{equation}
    \underbrace{\mathbb{E}_{x\sim\mathbb{P}_{\leq\tau}\left(x\right)}\left[\bm{1}_{y_\text{true}(x)=c}\right]}_{\text{frequency of correct predictions}}=\underbrace{\mathbb{E}_{x\sim\mathbb{P}_{\leq\tau}\left(x\right)}\left[\hat{f}\left(x\right)_c\right]}_{\text{predicted confidence}}
\end{equation}
where $\mathbb{P}_{\leq\tau}\left(x\right)$ is $\mathbb{P}\left(x\right)$ restricted to values of $x$ for which the confidence of the $c$ class $\hat{f}(x)_c$ is lower than or equal to $\tau$.

The calibration of a machine learning model can be assessed using different calibration metrics \citep{arrieta2022metrics}. A popular visual technique is to plot a reliability diagram (or calibration plot), by binning a set of instances (usually from a held-out calibration set) ranked by increasing confidence, then computing the mean confidence as well as the average observed accuracy in each bin. Improving the calibration of a machine learning model can be achieved by taking a fitted classifier and modifying its output confidences. In the \textbf{isotonic regression} calibration method \citep{zadrozny2001obtaining}, an isotonic regression model \citep{robertson1988order} is used to predict the actual mean accuracy from the mean confidence in each bin. At inference, the confidence score predicted by the classifier is then replaced by the output of the isotonic regression. In \textbf{Platt scaling} \citep{platt1999probabilistic}, the output of the predictor is transformed using the sigmoid function parameterized by scaling and bias scalar parameters. These parameters are learned so that the calibration metric is minimized on the calibration set.

\section{Calibration of model-probing detectors}
\label{sec:problem-statement}

\paragraph{Rationale - } The choice of the machine learning model used in model-probing detection methods is essential for their effectiveness. This model should be well suited to the specific learning task and at least somewhat robust against labeling noise \citep{george2024mislabeled}. In this article, we focus on another key aspect: calibration. %

Indeed, calibrated models exhibit some desirable properties. In addition to sometimes improving overall performance \citep[e.g., improved square loss in][]{caruana_good_proba_2005}, calibrating machine learning models has been reported to improve their robustness to noise \citep{gong2019large,zhang2023noisy}. In class-imbalanced datasets, calibration can also alleviate the effect of obtaining overconfident classifiers on the majority class at the cost of missing more examples of other classes \citep{huang_calib_imbalanced_2020}.
Furthermore, calibration often helps to obtain a better estimate of the true posterior $\mathbb{P}(Y|X=x)$, which can improve detection when using \emph{probes} based on confidence in the observed label $\mathbb{P}(Y=y_\text{observed}|X=x)$. Finally, calibration ensures that the predicted confidences are on similar scales across different models, which is especially useful for model-probing detectors that involve ensembling.

\paragraph{Proposed solution -} Starting from an existing mislabeled detection method, we propose adding an intermediate calibration step prior to probing. This is equivalent to replacing the original model with a calibrated model. Calibration only makes sense for models whose output is a confidence score interpreted as a probability (as opposed to e.g., a logit), which implies that our proposed solution only works with mislabeled detection methods that are based on the predicted confidence. In this case, the trust scores computed for each instance are modified, which can result in a re-ranking of training instances. For detectors using ensembling, all trained submodels are independently calibrated on the same held-out calibration set distinct from the training set (Figure \ref{fig:model_probing_calibration}).

In the experiments below, we employ isotonic regression \citep{zadrozny2001obtaining} and Platt scaling \citep{platt1999probabilistic} as our calibration methods. We verify empirically in Section \ref{sec:results} that the choice of the calibration algorithm does not matter much for our proposed approach to work, as long as calibration is improved on the training set. We emphasize that calibration has a low computational cost, vastly inferior to training the machine learning model(s).

\section{Experiments}
\label{sec:benchmark}

We evaluate our proposed solution by using a 3-stage pipeline:
$\text{\textbf{detection}} \Rightarrow \text{\textbf{filtering}} \Rightarrow \text{\textbf{training}}$. The test loss of the classifier trained from the filtered training dataset at the end of the pipeline is used as our performance metric. The performance of the \textbf{detection} method is thus measured as its ability to produce \textbf{filtered} datasets that contain examples that are useful for \textbf{training} a model: ideally, they would include as many correctly labeled examples as possible while being free from mislabeled ones.

More realistically, some mislabeled examples will likely be missed, and some correctly labeled examples will be incorrectly flagged as mislabeled. However, not all examples are equally important in regard to using them to learn a machine learning algorithm, such as in class-imbalanced datasets or fairness tasks with a minority group. In both of these cases, the minority class or group might disappear from the filtered dataset. %

\subsection{Datasets and labeling rules} We benchmark using the set of tasks and labeling noise from \cite{george2024mislabeled}. The benchmark is composed of 19 multiclass classification datasets on tabular and text data. These datasets are diverse in their number of samples, classes, features, and class balances, ranging from small and easy toy datasets to real-world ones (see Table \ref{tab:datasets}).

Each dataset is provided with labeling rules. A labeling rule is a heuristic function that takes an example's features and outputs a decision, which can be either a label or a non-response. For example, a labeling rule from the IMDB dataset states that: ``if a review contains the word \emph{pleasant}, it is a positive review''. Multiple labeling rules are provided by datasets, and the aggregation of all the labeling rules $\tilde{y}$ gives a noisy estimation of the true label $y_\text{true}$, especially in regions of the feature space not well covered by the labeling rules or when the labeling rules disagree. This type of label noise is representative of real-world label noise when datasets are annotated by many less knowledgeable experts or by automated data annotation rules (noisy not at random), even more so than the usual synthetic noise injected by flipping the label of randomly chosen examples from the training set (noisy completely at random).%

Each dataset comes with a predefined test set and, in some cases, a validation set. If no validation set is available, we reserve 20\% of the training set for validation. In our evaluation, the validation set is clean of label noise, as is the test set.

\begin{table}[t]
    \topcaption{Datasets used to benchmark detectors. Columns: dataset size $n$, number of features $d$, number of classes $C$, histogram of class priors $p(y)$, noise transition matrix $\mathbf{T}$, noise ratio $\rho$.}
    \centering
    \label{tab:datasets}
    \begin{tabular}{lrrrrrr}
        \toprule
        \textbf{Dataset}  & $n$   & $d$   & $C$ & $p(y)$                                                                                      & $\mathbf{T}$                                                                                           & $\rho$ \\
        \midrule
        hausa             & 2.92K & 750   & 5   & \parbox[c]{12pt}{\includegraphics[height=12pt]{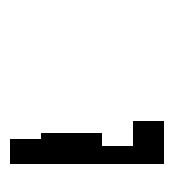}}             & \parbox[c]{12pt}{\includegraphics[height=12pt]{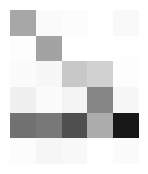}}             & 50\%   \\
        yoruba            & 1.91K & 539   & 7   & \parbox[c]{12pt}{\includegraphics[height=12pt]{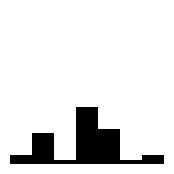}}            & \parbox[c]{12pt}{\includegraphics[height=12pt]{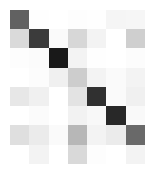}}            & 40\%   \\
        amazon            & 200K  & 3.68K & 2   & \parbox[c]{12pt}{\includegraphics[height=12pt]{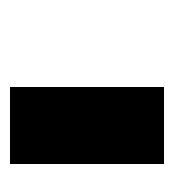}}            & \parbox[c]{12pt}{\includegraphics[height=12pt]{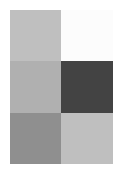}}            & 25\%   \\
        professor-teacher & 24.6K & 4.06K & 2   & \parbox[c]{12pt}{\includegraphics[height=12pt]{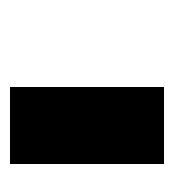}} & \parbox[c]{12pt}{\includegraphics[height=12pt]{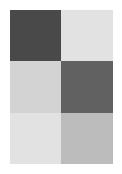}} & 18\%   \\
        bank-marketing    & 45.2K & 78    & 2   & \parbox[c]{12pt}{\includegraphics[height=12pt]{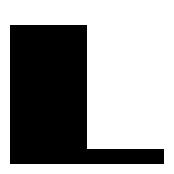}}    & \parbox[c]{12pt}{\includegraphics[height=12pt]{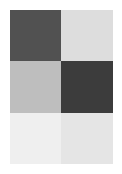}}    & 26\%   \\
        basketball        & 20.3K & 2.05K & 2   & \parbox[c]{12pt}{\includegraphics[height=12pt]{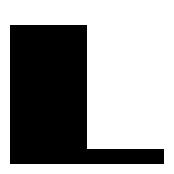}}        & \parbox[c]{12pt}{\includegraphics[height=12pt]{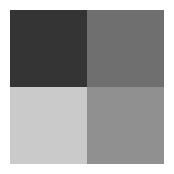}}        & 25\%   \\
        bioresponse       & 3.75K & 12.4K & 2   & \parbox[c]{12pt}{\includegraphics[height=12pt]{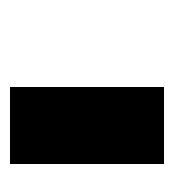}}       & \parbox[c]{12pt}{\includegraphics[height=12pt]{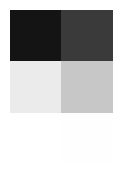}}       & 46\%   \\
        census            & 31.9K & 125   & 2   & \parbox[c]{12pt}{\includegraphics[height=12pt]{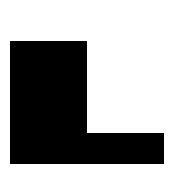}}            & \parbox[c]{12pt}{\includegraphics[height=12pt]{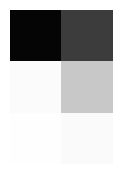}}            & 19\%   \\
        commercial        & 81.1K & 2.05K & 2   & \parbox[c]{12pt}{\includegraphics[height=12pt]{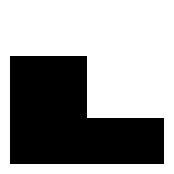}}        & \parbox[c]{12pt}{\includegraphics[height=12pt]{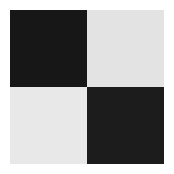}}        & 10\%   \\
        mushroom          & 8.12K & 128   & 2   & \parbox[c]{12pt}{\includegraphics[height=12pt]{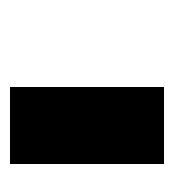}}          & \parbox[c]{12pt}{\includegraphics[height=12pt]{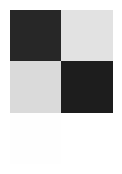}}          & 13\%   \\
        phishing          & 11.1K & 46    & 2   & \parbox[c]{12pt}{\includegraphics[height=12pt]{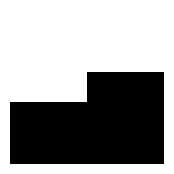}}          & \parbox[c]{12pt}{\includegraphics[height=12pt]{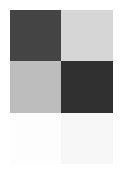}}          & 21\%   \\
        spambase          & 4.6K  & 57    & 2   & \parbox[c]{12pt}{\includegraphics[height=12pt]{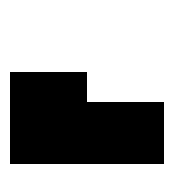}}          & \parbox[c]{12pt}{\includegraphics[height=12pt]{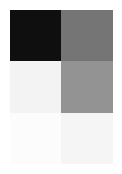}}          & 25\%   \\
        tennis            & 8.8K  & 2.05K & 2   & \parbox[c]{12pt}{\includegraphics[height=12pt]{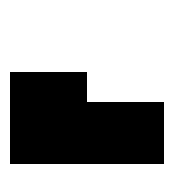}}            & \parbox[c]{12pt}{\includegraphics[height=12pt]{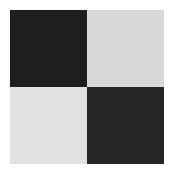}}            & 13\%   \\
        agnews            & 120K  & 3.36K & 4   & \parbox[c]{12pt}{\includegraphics[height=12pt]{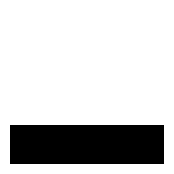}}            & \parbox[c]{12pt}{\includegraphics[height=12pt]{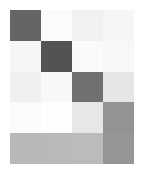}}            & 19\%   \\
        imdb              & 25K   & 12.1K & 2   & \parbox[c]{12pt}{\includegraphics[height=12pt]{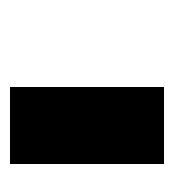}}              & \parbox[c]{12pt}{\includegraphics[height=12pt]{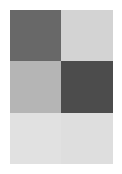}}              & 26\%   \\
        sms               & 5.57K & 1.37K & 2   & \parbox[c]{12pt}{\includegraphics[height=12pt]{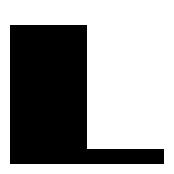}}               & \parbox[c]{12pt}{\includegraphics[height=12pt]{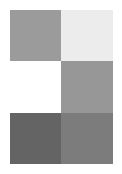}}               & 3\%    \\
        trec              & 6.03K & 946   & 6   & \parbox[c]{12pt}{\includegraphics[height=12pt]{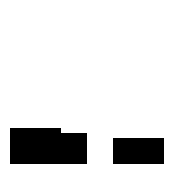}}              & \parbox[c]{12pt}{\includegraphics[height=12pt]{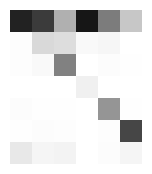}}              & 48\%   \\
        yelp              & 38K   & 5.35K & 2   & \parbox[c]{12pt}{\includegraphics[height=12pt]{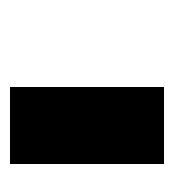}}              & \parbox[c]{12pt}{\includegraphics[height=12pt]{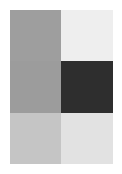}}              & 26\%   \\
        youtube           & 2.06K & 423   & 2   & \parbox[c]{12pt}{\includegraphics[height=12pt]{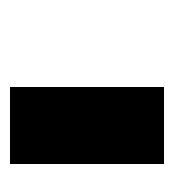}}           & \parbox[c]{12pt}{\includegraphics[height=12pt]{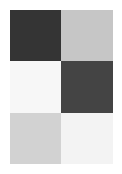}}           & 15\%   \\
        \bottomrule
    \end{tabular}
\end{table}

\subsection{Benchmark}

\paragraph{Detection methods} We benchmark multiple detection methods, including Area Under the Margin (AUM) \citep{pleiss_identifying_2020} which sums up confidence margins of a model during training, Small Loss \citep{amiri_spotting_2018, jiang2018mentornet_smallloss} which computes the log loss on each training instances of a model after training, Consensus Consistency \citep{jiang_characterizing_2021} which looks at disagreements between several models learned on bootstraps of the training set, and CleanLab \citep{northcutt2021confident} which evaluates the out-of-bag confidence of models learned on subsets of the training set (see Section \ref{sec:model_probing}). Detectors without modifications (\emph{calibration} or \emph{adjustment}) will be referred to as \emph{baselines}.

\paragraph{\emph{Adjust} method of \citet{kuan2022model}} As a competitor to \emph{calibration}, we test \emph{adjusting} the trust scores so that the average trust scores per class are centered and are then normalized to a valid probability distribution:
\begin{flalign}
    \label{eq:adj-center}
    \textit{centering:} &  & \tilde{p}_j(x) & = p_j(x) - \bar{p}_j + \underset{c}{\max}\, \bar{p}_c &
\end{flalign}
where $\bar{p}_j=\frac{1}{n_j}\sum_{i, y_i = j}p_j(x_i)$ is the average predicted $j$-th probability for all examples of class $j$, and $n_j$ is the number of examples of class $j$ in the dataset.
\begin{flalign}
    \label{eq:adj-norm}
    \textit{normalization:} &  & \tilde{p}_j(x) & = \frac{\tilde{p}_j(x)}{\sum_c \tilde{p}_c(x)} &
\end{flalign}

\paragraph{Calibration} We use isotonic regression (\emph{isotonic})  \citep{zadrozny2001obtaining} and Platt scaling (\emph{sigmoid}) \citep{platt1999probabilistic} to calibrate detectors (see Section \ref{sec:calibration}). When not specified, we reserve 50\% of the validation set as a held-out calibration set, and isotonic regression is used as a default.

\paragraph{References} We also tested four different scenarios to serve as a reference. The \emph{none} reference trains the final classifier on the noisy dataset without filtering; it serves as a worst-case scenario. The \emph{random} reference uses random trust scores to filter the noisy dataset. This reference is used because even when the trust scores provided by detection methods give no meaningful information, by chance a good subset of the dataset could be selected, containing few mislabeled examples, which would lead to a classifier with a good test loss. The \emph{silver} reference trains the final classifier on a perfectly filtered noisy dataset; it serves as a best-case scenario. The \emph{gold} reference trains the final classifier on the entire original dataset with true labels; it serves as a gold standard that filtering approaches should not reach.

We also use these references to normalize test losses across tasks as they are of varying difficulty. The test losses are linearly interpolated between 100 and 200 so that the \emph{none} reference gets 200 and the \emph{silver} reference gets 100.

\paragraph{Feature preprocessing and classifier} The base machine learning model used in this study is a linear classifier (LC) trained with stochastic gradient descent to minimize the log loss. For text datasets, we compute the TF-IDF features, which are $\ell_2$ normalized. For tabular datasets, categorical features are one-hot encoded, and numerical features are normalized; then, non-linear random features are generated with Random Fourier Features (RFF) \citep{NIPS2007_013a006f}. We generate these features as classes might not be linearly separable in the original feature space.

\paragraph{Hyperparameter optimization} To select hyperparameter values, we perform a random search per detector and task for the learning rate and the $\ell_2$ regularization of the detector's base model and final classifier (see Table \ref{tab:hyperparameters}). For each sampled set of hyperparameters, we test filtering out of the train set the examples with the 10\%, 20\%, ..., and 90\% lowest trust scores. We then select the pipeline that yielded the final classifier obtaining the minimum log loss on a clean validation set out of the 1440 pipelines generated through the random search. This classifier is then evaluated on a test set, and the results are reported in the next section.

\begin{table}[!h]
    \topcaption{Table of hyperparameters}
    \centering
    \begin{tabular}{rll}
        \toprule
                                                      & \textbf{Hyperparameter} & \textbf{Search space}                                    \\
        \midrule
        \multirow{2}{*}{\rotatebox[origin=c]{90}{RFF}}
                                                      & Monte Carlo samples     & 1000                                                     \\
                                                      & kernel bandwith         & inverse of dataset's variance (see \citep{scikit-learn}) \\
        \midrule
        \multirow{2}{*}{\rotatebox[origin=c]{90}{LC}} & $\ell_2$ regularization & log-uniform $[1e^{-5}, 1e^{-1}]$                         \\
                                                      & learning rate           & log-uniform $[1e^{-3}, 1]$                               \\
        \bottomrule
    \end{tabular}
    \label{tab:hyperparameters}
\end{table}

\section{Results}
\label{sec:results}

\paragraph{Does calibration help figuring out hard examples from mislabeled ones?}

One of our initial motivations for calibrating model-probing detectors was to help them differentiate mislabeled examples from genuine but hard examples where the base model tends to be under-confident, such as examples from rarer subspaces of the input space. As a proxy for such a setup with easy and hard examples, we consider \emph{imbalanced} classification datasets, where the minority class examples are more difficult to correctly classify. We look at how much detectors consider examples from the minority classes as untrusted just because they are from the minority. We consider minority classes the classes with a prior less than $1/C$ (random prior), generalizing the definition for multi-class datasets. We measured the number of examples from these minority classes removed as we filtered out examples from the lowest to the highest trust scores.

\begin{figure}[h]
\centering
\includegraphics[width=0.75\linewidth]{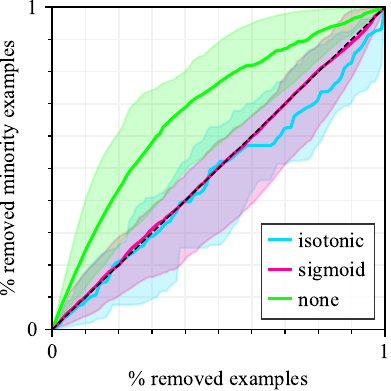}
\bottomcaption{Evolution of the number of removed examples when filtering instances from the least to the most trusted ones on the x-axis, with the number of removed examples from the minority classes (classes with priors less than $1/C$) on the y-axis. We plot the median ratio over all datasets alongside the 25\% and 75\% percentiles for detectors without calibration (none) and detectors with Platt scaling (sigmoid) and isotonic regression. Detectors without calibration have the tendency to consider minority examples as untrusted, whereas calibrated detectors do not.}
\label{fig:minority_removed}
\end{figure}

\begin{figure*}[t]
    \centering    
    \includegraphics[width=0.8\linewidth]{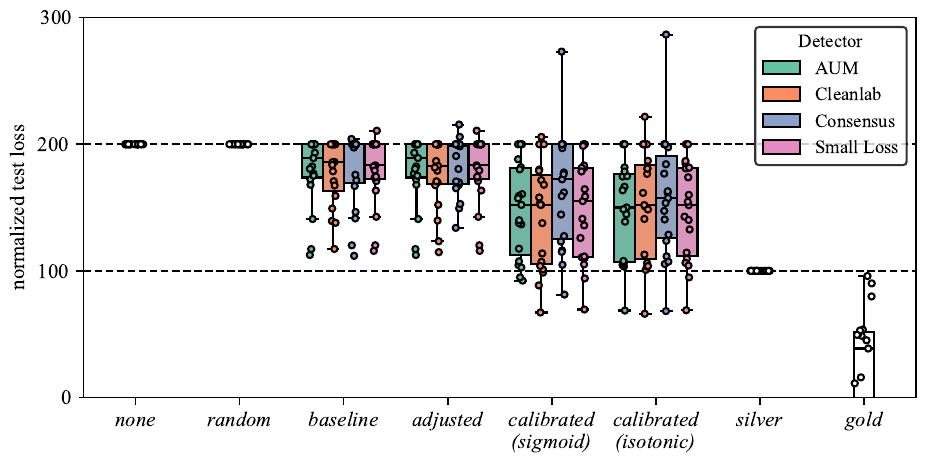}
    \bottomcaption{Distribution (boxplot) of the normalized (base 100, \emph{silver} = training on correctly labeled examples only, base 200, \emph{none} = training on all examples including mislabeled ones) test loss (lower is better) of the final classifier after the 3-stages pipeline of varying detectors over all datasets (the circles $\bullet$). Classifiers are calibrated using a clean calibration set. Calibrated detectors are significantly better than their adjusted or baseline counterparts.}
    \label{fig:plugins_clean_valid}
\end{figure*}

Figure \ref{fig:minority_removed} shows that for uncalibrated detectors (the green curve), the ratio of removed examples from minority classes grows faster than the overall ratio of removed examples. This indicates that examples from minority classes are assigned lower trust scores on average compared to those from majority classes. For calibrated detectors with both isotonic regression and Platt scaling (blue and purple curves), both ratio grows at the same speed. This suggests that calibration helps detectors balance trust scores between minority and majority examples in our benchmark. However, adjusting probabilities did not help in this case (see Figure \ref{fig:minority_removed_adjusted} in Appendix \ref{sec:additional-experiments}).

\paragraph{Does the ranking provided by a calibrated detection method allow training better classifiers?} 

For each detection method, we compare their performance metric (normalized test loss of the classifier obtained after the 3-stage pipeline) with the performance metrics for their \textit{adjusted} and \textit{calibrated} variants. In our experiments (Figure \ref{fig:plugins_clean_valid}), we observe that the test loss is lower when using \textit{calibrated} detectors and that \textit{adjusted} has little to no impact on detectors' performances.

To measure the significance of this improvement, we conduct hypothesis testing with a Wilcoxon signed-rank test \citep{conover1999practical} to assess whether the observed differences in test loss and balanced accuracy between the detectors with and without calibration are statistically significant. Tables \ref{tab:results-logl} and \ref{tab:results-bacc} in Appendix \ref{sec:detailed-results} show the results of these tests for both sigmoid and isotonic calibration at a significance level of 95\%. For all detectors and calibration methods, the improvement provided by the calibration step is significant. Tables \ref{tab:results-logl} and \ref{tab:results-bacc} in Appendix \ref{sec:detailed-results}, also provide detailed per-dataset results.

\begin{figure}[h]
\centering
\includegraphics[width=0.8\linewidth]{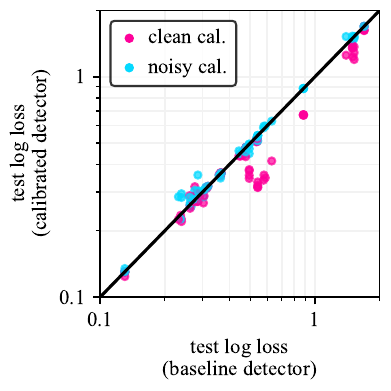}
\bottomcaption{For each detector/dataset pair (a circle), we compare the test loss between a baseline detector on the x-axis, with the test loss of the same detector calibrated on a clean calibration set \textcolor{xkcdNeonPink}{$\bullet$} and on a noisy calibration set \textcolor{xkcdNeonBlue}{$\bullet$} on the y-axis. A clean calibration set is often the most efficient (\textcolor{xkcdNeonPink}{$\bullet$} are mainly below the $y=x$ line). A noisy calibration set, although less efficient, is not significantly worse than the baseline (\textcolor{xkcdNeonBlue}{$\bullet$} are equally distributed around the $y=x$ line).}
\label{fig:clean_vs_noisy}
\end{figure}

\paragraph{Do we need to calibrate on a clean calibration set?}
To assess the robustness of our approach, we evaluate how the detectors perform when the calibration set also contains mislabeled examples, with a labeling process corrupted in the same manner as the labeling process of the training set. Figure \ref{fig:clean_vs_noisy} illustrates that the performance of detectors \emph{calibrated} with a clean calibration set is better than detectors \emph{calibrated} with a noisy calibration set. However, even when calibration is performed on a noisy set, performance is not significantly worse than the \emph{baseline} detector one. In other words, the calibration step seems to be consistently beneficial, regardless of the label noise present in the calibration set.

\paragraph{How big should the calibration set be?}

\begin{figure*}[t]
\centering
\includegraphics[width=0.8\linewidth]{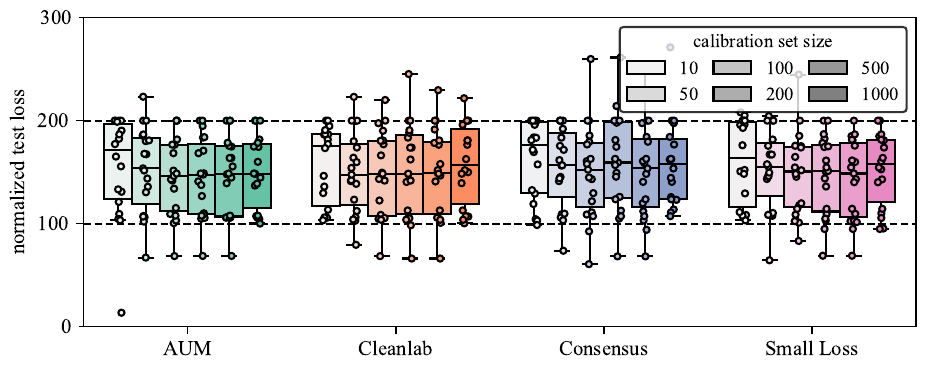}
\bottomcaption{Distribution (boxplot) of the normalized (base 100, \emph{silver} = training on correctly labeled examples only, base 200, \emph{none} = training on all examples including mislabeled ones) test loss (lower is better) of the final classifier after the 3-stages pipeline of varying calibrated detectors over all datasets (the circles $\bullet$). Classifiers are calibrated using a clean calibration set with varying sizes, from 10 samples (white boxplot) to 1000 samples (colorful boxplot). There are diminishing returns in adding more calibration samples after 100 samples.}
\label{fig:size_calib}
\end{figure*}

We also evaluate the robustness of our approach to the size of the calibration set. Even though we showed that having a clean calibration set is not required for the efficiency of our approach, we want to emphasize that as few as dozens of samples are enough to calibrate classifiers properly. In Figure \ref{fig:size_calib}, we compare the performance of the \emph{calibrated} detector when adding increasingly more samples to the calibration set, from a dozen samples to thousands (note that this number is capped for the smallest datasets). Calibrating detectors starts to be beneficial after dozens of calibration samples, and starts to plateau in the hundreds. That can be explained by the sample efficiency of isotonic regression, needing just a few samples to improve calibration \citep{dai2020bias}.

\begin{figure}[h]
\centering
\includegraphics[width=0.8\linewidth]{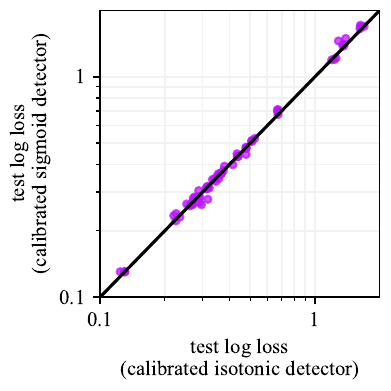}
\bottomcaption{For each detector/dataset pair (a circle), we compare the test loss between a detector calibrated with isotonic regression on the x-axis with the test loss of the same detector calibrated with Platt scaling (sigmoid) on the y-axis. The performance of both calibration methods is highly correlated, suggesting that our approach is not dependent on the choice of the calibration method.}
\label{fig:corr_calibration}
\end{figure}

\paragraph{How much impact has the calibration method?}
\label{sec:impact}
Finally, we assess the impact of the choice of the calibration method on our approach with: isotonic regression and Platt scaling. Figure \ref{fig:plugins_clean_valid} demonstrates that both methods significantly improve the quality of detectors. Furthermore, Figure \ref{fig:corr_calibration} illustrates a strong correlation between the improvements achieved by both calibration methods. Even though on average both calibration methods worked equally well on our benchmark, we suggest practitioners to pick the calibration strategy best suited to their model and dataset. For instance, temperature scaling \citep{guo2017calibration} is particularly effective for deep networks. %

\section{Conclusion}

In this paper, we introduced calibration as a simple add-on to model-probing mislabeled detection techniques. We motivated it as a way to obtain a better and more reliable underlying model, particularly in balancing trust scores between minority and majority classes. Through extensive experiments on a large benchmark of realistic noisy datasets, generated using weak labeling rules, we demonstrate that our technique significantly improves the test loss of the final classifier obtained in fully automated $\text{\textbf{detection}} \Rightarrow \text{\textbf{filtering}} \Rightarrow \text{\textbf{training}}$ pipelines. Furthermore, we showed that our proposed methodology is robust to label noise in the calibration set and is effective even with small calibration sets.

Our findings suggest that calibration should be considered a standard component in mislabeled detection pipelines, particularly in real-world scenarios with class imbalance or underrepresented subgroups. This simple modification provides consistent improvements across multiple detection methods and datasets, making it an accessible and practical solution for industrial applications dealing with mislabeled examples.

\bibliography{biblio.bib}

\clearpage
\appendix
\onecolumn

\section{Details on class-imbalanced motivating examples}\label{sec:motivating_examples}

\paragraph{Toy dataset, 2-moons:} As a motivating example, we examine the impact of adding a calibration stage to the AUM detector on the 2-moons dataset from scikit-learn \citep{scikit-learn}, a binary classification task where instances lie in a 2d feature space. In this experiment, we sample 100 instances in a 10\%/90\% class-imbalanced fashion. To introduce some mislabeling, we flip the labels of five examples from each class. The model used is an SGDClassifier with an RBF kernel from scikit-learn. We apply the AUM detection method, both with and without calibration, to obtain trust scores for each instance and  compare the top 10 less trusted instances in Figure \ref{fig:two_moons_pie}.

We observe that the strongest change between the \textbf{baseline} and the \textbf{calibrated} detector affects the minority class (in green).  Without calibration, the AUM method tends to misclassify all examples from the minority class as mislabeled due to the classifier's overconfidence in the majority class and underconfidence in the minority class. Calibrating the classifier helps alleviate this issue, and the mislabeled examples from the minority class (observed as examples from the majority class, the $\star$ in the bottom right corner) are correctly detected, and retain correctly labeled minority class examples that were previously discarded.

\paragraph{SMS dataset: } We decline the previous example on a real-world dataset. The SMS dataset \citep{sms_spam_filtering_2011} is a spam detection classification dataset. The corpus contains English SMS gathered from multiple sources, and the task is to classify legitimate messages from spams. The dataset is highly imbalanced, composed of 90\% of legitimate messages. Instead of using the original provided labels, we instead use key-word based labeling rules \citep{Awasthi2020Learning} to mimic the labels produced by automated or decentralized annotations methods used to quickly annotate web-scale datasets. We also apply the AUM detection method with and without calibration to detect suspicious labels generated from the key-word rules.

Table \ref{table:sms} shows the top 5 examples with the lowest trust scores for the \textbf{baseline} AUM and the \textbf{calibrated} AUM. In this experiment, the \textbf{baseline} AUM method only selects instances from the minority class, which amplifies class imbalance in the trusted subset after filtering. By contrast, adding an intermediate calibration step allows the detection method to parsimoniously select examples from both classes. 

This behavior is especially interesting for spam or fraud detection tasks, where missing spam or fraud in the annotation process can be costly, as they are rarer and more important examples to focus on than missing legitimate examples.  

\begin{table}[h]
\centering
\begin{minipage}{.5\textwidth}
\topcaption{
    Top 5 most suspicious examples found in the SMS dataset by the Area Under the Margin  (AUM) detector with a non-calibrated (\textbf{baseline}) model and a \textbf{calibrated} model. The \textbf{baseline} detector frequently assumes that SMS labeled as spam (the minority class) are mislabeled, meanwhile the \textbf{calibrated} detector returns spams that were not caught yet as they were still labeled as legitimate messages.}
\centering
\begin{tabular}{lp{0.5\textwidth}cc}
    \toprule
                                                                    & \multirow{2}{*}{SMS}                                                                                                                                                          & \multicolumn{2}{c}{Label}          \\
                                                                    &                                                                                                                                                                               & Noisy                     & True   \\
    \midrule
    \multirow{16}{*}{\rotatebox[origin=c]{90}{\textbf{Baseline}}}   & ok i've sent u da latest version of da project.                                                                                                                               & spam                      & legit. \\
                                                                    & freemsg hey there darling it's been 3 week's now and no word back! i'd like some fun you up for it still? tb ok! xxx std chgs to send, ??1.50 to rcv                          & spam                      & spam   \\
                                                                    & i don't know u and u don't know me. send chat to 86688 now and let's find each other! only 150p/msg rcvd. hg/suite342/2lands/row/w1j6hl ldn. 18 years or over.                & spam                      & spam   \\
                                                                    & i don't know u and u don't know me. send chat to 86688 now and let's find each other! only 150p/msg rcvd. hg/suite342/2lands/row/w1j6hl ldn. 18 years or over.                & spam                      & spam   \\
                                                                    & aiyo cos i sms ?\_ then ?\_ neva reply so i wait 4 ?\_ to reply lar. i tot ?\_ havent finish ur lab wat.                                                                      & spam                      & legit. \\
    \midrule
    \multirow{13}{*}{\rotatebox[origin=c]{90}{\textbf{Calibrated}}} & do you want a new nokia 3510i colour phone delivered tomorrow? with 200 free minutes to any mobile + 100 free text + free camcorder reply or call 8000930705                  & legit.                    & spam   \\
                                                                    & 44 7732584351, do you want a new nokia 3510i colour phone deliveredtomorrow? with 300 free minutes to any mobile + 100 free texts + free camcorder reply or call 08000930705. & legit.                    & spam   \\
                                                                    & if you were/are free i can give. otherwise nalla adi entey nattil kittum                                                                                                      & legit.                    & legit. \\
                                                                    & are you free now?can i call now?                                                                                                                                              & legit.                    & legit. \\
                                                                    & ok i've sent u da latest version of da project.                                                                                                                               & spam                      & legit. \\
    \bottomrule
\end{tabular}
\label{table:sms}

\end{minipage}
\end{table}

\clearpage

\section{Detailed benchmark results}\label{sec:detailed-results}

Detailed benchmark results are provided below in Tables \ref{tab:results-logl} and \ref{tab:results-bacc} for the test loss and balanced accuracy for each detector on each dataset on top of the \emph{silver} and \emph{none} reference. For each detector and dataset, we \textbf{bolded} the result of the best add-on. For example, in Table \ref{tab:results-logl} for \emph{agnews} and the AUM detector, both isotonic (\emph{iso.}) and sigmoid (\emph{sig.}) calibration got the best log-loss of \textbf{0.51}.

We also provided the number of wins, draws and losses of \emph{calibrated} and \emph{adjusted} detectors against the \emph{baseline} (no calibration) detector. If the number of wins, draws and losses is \underline{\textbf{underlined}}, the difference between the two detectors is significant. We conducted hypothesis testing with a Wilcoxon signed-rank test \citep{conover1999practical} at a significance level of 95\%.

\begin{table*}[h]
    \centering
    \topcaption{Test loss of the best pipelines for each detector and add-on for all datasets. The results in \textbf{bold} are the results obtained by the best add-on for each detector on each dataset. The results of the silver and none references are in \textit{italic}.}
    \label{tab:results-logl}
    \resizebox{\linewidth}{!}{
        \begin{tabular}{lrrrrrrrrrrrrrrrrrr}
            \toprule
            \multirow{2.5}{*}{\textbf{Dataset}} & \multirow{2.5}{*}{silver} & \multicolumn{4}{c}{AUM} & \multicolumn{4}{c}{Cleanlab} & \multicolumn{4}{c}{Consensus} & \multicolumn{4}{c}{Small Loss} & \multirow{2.5}{*}{none}                                                                                                                                                                                                                                                                                                    \\
            \cmidrule(lr){3-6} \cmidrule(lr){7-10} \cmidrule(lr){11-14} \cmidrule(lr){15-18}
                                                &                           & \textit{base.}          & \textit{adj.}                & \textit{iso.}                 & \textit{sig.}                  & \textit{base.}          & \textit{adj.}   & \textit{iso.}                & \textit{sig.}                & \textit{base.} & \textit{adj.}   & \textit{iso.}                & \textit{sig.}                & \textit{base.} & \textit{adj.}    & \textit{iso.}                & \textit{sig.}                &               \\
            \midrule
            agnews                              & \itshape 0.43             & 0.53                    & 0.53                         & \bfseries 0.51                & \bfseries 0.51                 & 0.54                    & 0.54            & \bfseries 0.51               & \bfseries 0.51               & 0.54           & 0.53            & \bfseries 0.52               & \bfseries 0.52               & 0.54           & 0.54             & \bfseries 0.51               & \bfseries 0.51               & \itshape 0.58 \\
            amazon                              & \itshape 0.33             & 0.58                    & 0.58                         & \bfseries 0.34                & \bfseries 0.34                 & 0.59                    & 0.59            & \bfseries 0.35               & \bfseries 0.35               & 0.63           & 0.63            & 0.41                         & \bfseries 0.40               & 0.58           & 0.58             & \bfseries 0.36               & \bfseries 0.36               & \itshape 0.63 \\
            bank-marketing                      & \itshape 0.27             & 0.28                    & 0.28                         & \bfseries 0.27                & \bfseries 0.27                 & 0.30                    & 0.29            & \bfseries 0.27               & \bfseries 0.27               & 0.29           & 0.33            & \bfseries 0.27               & 0.28                         & 0.29           & 0.29             & \bfseries 0.27               & 0.28                         & \itshape 0.36 \\
            basketball                          & \itshape 0.25             & 0.29                    & 0.29                         & 0.29                          & \bfseries 0.27                 & \bfseries 0.28          & 0.29            & 0.32                         & \bfseries 0.28               & 0.27           & 0.28            & 0.27                         & \bfseries 0.26               & 0.28           & 0.28             & 0.30                         & \bfseries 0.26               & \itshape 0.30 \\
            bioresponse                         & \itshape 0.72             & 0.88                    & 0.88                         & \bfseries 0.67                & 0.71                           & 0.88                    & 0.88            & \bfseries 0.67               & 0.70                         & 0.88           & 0.78            & \bfseries 0.67               & 0.69                         & 0.88           & 0.88             & \bfseries 0.67               & \bfseries 0.67               & \itshape 0.88 \\
            census                              & \itshape 0.45             & \bfseries 0.48          & \bfseries 0.48               & \bfseries 0.48                & \bfseries 0.48                 & 0.48                    & 0.48            & 0.48                         & \bfseries 0.44               & \bfseries 0.48 & \bfseries 0.48  & \bfseries 0.48               & \bfseries 0.48               & 0.48           & 0.48             & 0.48                         & \bfseries 0.47               & \itshape 0.48 \\
            commercial                          & \itshape 0.25             & \bfseries 0.26          & \bfseries 0.26               & \bfseries 0.26                & \bfseries 0.26                 & 0.27                    & \bfseries 0.26  & 0.27                         & \bfseries 0.26               & \bfseries 0.26 & 0.27            & 0.29                         & 0.29                         & 0.26           & 0.26             & \bfseries 0.25               & 0.26                         & \itshape 0.27 \\
            hausa                               & \itshape 1.21             & 1.50                    & 1.50                         & 1.23                          & \bfseries 1.20                 & 1.40                    & 1.37            & 1.25                         & \bfseries 1.22               & 1.52           & 1.52            & \bfseries 1.29               & 1.45                         & 1.52           & 1.52             & 1.20                         & \bfseries 1.19               & \itshape 1.52 \\
            imdb                                & \itshape 0.38             & 0.46                    & 0.46                         & \bfseries 0.44                & \bfseries 0.44                 & 0.44                    & 0.45            & 0.44                         & \bfseries 0.43               & 0.48           & 0.47            & \bfseries 0.43               & 0.45                         & 0.45           & 0.45             & 0.44                         & \bfseries 0.43               & \itshape 0.47 \\
            mushroom                            & \itshape 0.18             & \bfseries 0.28          & \bfseries 0.28               & \bfseries 0.28                & \bfseries 0.28                 & 0.29                    & \bfseries 0.28  & \bfseries 0.28               & \bfseries 0.28               & 0.30           & 0.31            & \bfseries 0.29               & 0.30                         & 0.29           & 0.29             & 0.29                         & \bfseries 0.28               & \itshape 0.31 \\
            phishing                            & \itshape 0.23             & 0.37                    & 0.37                         & 0.37                          & \bfseries 0.36                 & \bfseries 0.36          & 0.37            & 0.37                         & \bfseries 0.36               & \bfseries 0.36 & 0.37            & \bfseries 0.36               & \bfseries 0.36               & 0.36           & 0.36             & \bfseries 0.35               & 0.36                         & \itshape 0.40 \\
            professor-teacher                   & \itshape 0.28             & \bfseries 0.31          & \bfseries 0.31               & \bfseries 0.31                & \bfseries 0.31                 & \bfseries 0.31          & \bfseries 0.31  & \bfseries 0.31               & \bfseries 0.31               & \bfseries 0.32 & \bfseries 0.32  & \bfseries 0.32               & \bfseries 0.32               & \bfseries 0.31 & \bfseries 0.31   & \bfseries 0.31               & \bfseries 0.31               & \itshape 0.36 \\
            sms                                 & \itshape 0.10             & \bfseries 0.13          & \bfseries 0.13               & \bfseries 0.13                & \bfseries 0.13                 & \bfseries 0.13          & \bfseries 0.13  & \bfseries 0.13               & \bfseries 0.13               & \bfseries 0.13 & \bfseries 0.13  & \bfseries 0.13               & \bfseries 0.13               & 0.13           & 0.13             & \bfseries 0.12               & 0.13                         & \itshape 0.13 \\
            spambase                            & \itshape 0.25             & 0.49                    & 0.49                         & 0.36                          & \bfseries 0.35                 & 0.49                    & 0.49            & \bfseries 0.37               & 0.38                         & 0.49           & 0.42            & \bfseries 0.38               & 0.39                         & 0.49           & 0.49             & 0.35                         & \bfseries 0.33               & \itshape 0.49 \\
            tennis                              & \itshape 0.36             & \bfseries 0.37          & \bfseries 0.37               & \bfseries 0.37                & \bfseries 0.37                 & \bfseries 0.37          & \bfseries 0.37  & \bfseries 0.37               & \bfseries 0.37               & \bfseries 0.37 & \bfseries 0.37  & \bfseries 0.37               & \bfseries 0.37               & \bfseries 0.37 & \bfseries 0.37   & \bfseries 0.37               & \bfseries 0.37               & \itshape 0.37 \\
            trec                                & \itshape 1.35             & 1.69                    & 1.69                         & \bfseries 1.62                & 1.65                           & 1.69                    & 1.69            & \bfseries 1.63               & 1.71                         & \bfseries 1.69 & \bfseries 1.69  & \bfseries 1.69               & \bfseries 1.69               & 1.69           & 1.69             & \bfseries 1.63               & 1.69                         & \itshape 1.69 \\
            yelp                                & \itshape 0.31             & 0.54                    & 0.54                         & 0.32                          & \bfseries 0.31                 & 0.54                    & 0.54            & \bfseries 0.31               & 0.32                         & 0.54           & 0.52            & \bfseries 0.33               & 0.34                         & 0.54           & 0.54             & \bfseries 0.32               & \bfseries 0.32               & \itshape 0.54 \\
            yoruba                              & \itshape 1.22             & 1.49                    & 1.49                         & \bfseries 1.35                & 1.37                           & 1.49                    & 1.49            & \bfseries 1.35               & 1.40                         & 1.49           & 1.49            & \bfseries 1.39               & 1.49                         & 1.52           & 1.52             & \bfseries 1.37               & 1.39                         & \itshape 1.49 \\
            youtube                             & \itshape 0.21             & 0.24                    & 0.24                         & \bfseries 0.23                & 0.24                           & 0.24                    & 0.24            & \bfseries 0.22               & 0.23                         & 0.23           & 0.29            & 0.23                         & \bfseries 0.22               & 0.24           & 0.24             & 0.24                         & \bfseries 0.23               & \itshape 0.35 \\
            \midrule
            wins/draws/losses                   & -                         & -                       & \bfseries 0/19/0             & \bfseries \underline{15/3/1}  & \bfseries \underline{14/3/2}   & -                       & \bfseries 6/8/5 & \bfseries \underline{12/3/4} & \bfseries \underline{14/2/3} & -              & \bfseries 5/6/8 & \bfseries \underline{13/3/3} & \bfseries \underline{12/5/2} & -              & \bfseries 0/19/0 & \bfseries \underline{14/2/3} & \bfseries \underline{16/2/1} & -             \\
            \bottomrule
        \end{tabular}
    }
\end{table*}

\begin{table*}[h]
    \centering
    \topcaption{Test balanced accuracy of the best pipelines for each detector and add-on for all datasets. The results in \textbf{bold} are the results obtained by the best add-on for each detector on each dataset. The results of the silver and none references are in \textit{italic}.}
    \label{tab:results-bacc}
    \resizebox{\linewidth}{!}{
        \begin{tabular}{lrrrrrrrrrrrrrrrrrr}
            \toprule
            \multirow{2.5}{*}{\textbf{Dataset}} & \multirow{2.5}{*}{silver} & \multicolumn{4}{c}{AUM} & \multicolumn{4}{c}{Cleanlab} & \multicolumn{4}{c}{Consensus} & \multicolumn{4}{c}{Small Loss} & \multirow{2.5}{*}{none}                                                                                                                                                                                                                                                                                                    \\
            \cmidrule(lr){3-6} \cmidrule(lr){7-10} \cmidrule(lr){11-14} \cmidrule(lr){15-18}
                                                &                           & \textit{base.}          & \textit{adj.}                & \textit{iso.}                 & \textit{sig.}                  & \textit{base.}          & \textit{adj.}   & \textit{iso.}                & \textit{sig.}                & \textit{base.} & \textit{adj.}    & \textit{iso.}                & \textit{sig.}               & \textit{base.} & \textit{adj.}    & \textit{iso.}                & \textit{sig.}                &               \\
            \midrule
            agnews                              & \itshape 0.86             & 0.82                    & 0.82                         & \bfseries 0.83                & \bfseries 0.83                 & 0.82                    & 0.82            & \bfseries 0.83               & \bfseries 0.83               & 0.81           & 0.82             & \bfseries 0.83               & 0.82                        & 0.82           & 0.82             & \bfseries 0.83               & \bfseries 0.83               & \itshape 0.81 \\
            amazon                              & \itshape 0.86             & 0.67                    & 0.67                         & 0.85                          & \bfseries 0.86                 & 0.67                    & 0.67            & \bfseries 0.85               & \bfseries 0.85               & 0.67           & 0.69             & \bfseries 0.84               & \bfseries 0.84              & 0.67           & 0.67             & \bfseries 0.85               & \bfseries 0.85               & \itshape 0.67 \\
            bank-marketing                      & \itshape 0.80             & \bfseries 0.79          & \bfseries 0.79               & \bfseries 0.79                & \bfseries 0.79                 & \bfseries 0.79          & \bfseries 0.79  & \bfseries 0.79               & \bfseries 0.79               & \bfseries 0.79 & \bfseries 0.79   & \bfseries 0.79               & \bfseries 0.79              & \bfseries 0.79 & \bfseries 0.79   & \bfseries 0.79               & \bfseries 0.79               & \itshape 0.79 \\
            basketball                          & \itshape 0.83             & 0.64                    & 0.64                         & \bfseries 0.67                & 0.65                           & 0.63                    & 0.67            & \bfseries 0.74               & 0.61                         & 0.61           & \bfseries 0.68   & 0.61                         & 0.61                        & \bfseries 0.63 & \bfseries 0.63   & 0.62                         & 0.60                         & \itshape 0.61 \\
            bioresponse                         & \itshape 0.65             & 0.60                    & 0.60                         & 0.64                          & \bfseries 0.65                 & 0.60                    & 0.60            & 0.63                         & \bfseries 0.64               & 0.60           & 0.63             & \bfseries 0.64               & \bfseries 0.64              & 0.60           & 0.60             & \bfseries 0.64               & \bfseries 0.64               & \itshape 0.60 \\
            census                              & \itshape 0.62             & \bfseries 0.63          & \bfseries 0.63               & \bfseries 0.63                & \bfseries 0.63                 & 0.63                    & 0.63            & 0.63                         & \bfseries 0.65               & \bfseries 0.63 & \bfseries 0.63   & \bfseries 0.63               & \bfseries 0.63              & 0.63           & 0.63             & \bfseries 0.64               & \bfseries 0.64               & \itshape 0.63 \\
            commercial                          & \itshape 0.90             & 0.89                    & 0.89                         & \bfseries 0.90                & \bfseries 0.90                 & 0.89                    & 0.89            & \bfseries 0.90               & 0.89                         & \bfseries 0.90 & \bfseries 0.90   & 0.89                         & 0.89                        & \bfseries 0.89 & \bfseries 0.89   & \bfseries 0.89               & \bfseries 0.89               & \itshape 0.89 \\
            hausa                               & \itshape 0.48             & 0.40                    & 0.40                         & 0.50                          & \bfseries 0.53                 & 0.40                    & 0.42            & 0.46                         & \bfseries 0.50               & 0.40           & 0.40             & \bfseries 0.42               & 0.41                        & 0.40           & 0.40             & 0.48                         & \bfseries 0.52               & \itshape 0.40 \\
            imdb                                & \itshape 0.83             & 0.78                    & 0.78                         & \bfseries 0.81                & \bfseries 0.81                 & \bfseries 0.80          & \bfseries 0.80  & \bfseries 0.80               & \bfseries 0.80               & 0.77           & 0.77             & \bfseries 0.80               & 0.79                        & 0.79           & 0.79             & \bfseries 0.80               & \bfseries 0.80               & \itshape 0.77 \\
            mushroom                            & \itshape 0.93             & \bfseries 0.90          & \bfseries 0.90               & 0.89                          & \bfseries 0.90                 & 0.87                    & 0.87            & \bfseries 0.90               & \bfseries 0.90               & 0.87           & 0.87             & \bfseries 0.91               & 0.87                        & \bfseries 0.90 & \bfseries 0.90   & 0.89                         & \bfseries 0.90               & \itshape 0.87 \\
            phishing                            & \itshape 0.91             & \bfseries 0.84          & \bfseries 0.84               & 0.83                          & 0.83                           & 0.84                    & 0.83            & \bfseries 0.85               & 0.83                         & 0.84           & 0.83             & \bfseries 0.85               & 0.84                        & 0.83           & 0.83             & \bfseries 0.84               & 0.83                         & \itshape 0.81 \\
            professor-teacher                   & \itshape 0.89             & \bfseries 0.88          & \bfseries 0.88               & \bfseries 0.88                & \bfseries 0.88                 & \bfseries 0.88          & \bfseries 0.88  & \bfseries 0.88               & \bfseries 0.88               & \bfseries 0.88 & \bfseries 0.88   & \bfseries 0.88               & \bfseries 0.88              & \bfseries 0.88 & \bfseries 0.88   & \bfseries 0.88               & \bfseries 0.88               & \itshape 0.87 \\
            sms                                 & \itshape 0.87             & \bfseries 0.84          & \bfseries 0.84               & 0.83                          & \bfseries 0.84                 & 0.84                    & 0.84            & 0.84                         & \bfseries 0.89               & 0.84           & \bfseries 0.85   & 0.83                         & 0.84                        & 0.84           & 0.84             & 0.86                         & \bfseries 0.90               & \itshape 0.84 \\
            spambase                            & \itshape 0.88             & 0.69                    & 0.69                         & \bfseries 0.82                & 0.80                           & 0.69                    & 0.69            & \bfseries 0.81               & \bfseries 0.81               & 0.69           & 0.76             & \bfseries 0.79               & \bfseries 0.79              & 0.69           & 0.69             & \bfseries 0.85               & \bfseries 0.85               & \itshape 0.69 \\
            tennis                              & \itshape 0.88             & \bfseries 0.88          & \bfseries 0.88               & \bfseries 0.88                & \bfseries 0.88                 & \bfseries 0.88          & \bfseries 0.88  & \bfseries 0.88               & \bfseries 0.88               & \bfseries 0.88 & \bfseries 0.88   & \bfseries 0.88               & \bfseries 0.88              & \bfseries 0.88 & \bfseries 0.88   & \bfseries 0.88               & \bfseries 0.88               & \itshape 0.88 \\
            trec                                & \itshape 0.48             & 0.31                    & 0.31                         & 0.38                          & \bfseries 0.39                 & 0.31                    & 0.31            & \bfseries 0.36               & 0.34                         & 0.31           & \bfseries 0.38   & 0.34                         & 0.37                        & 0.31           & 0.31             & 0.27                         & \bfseries 0.40               & \itshape 0.31 \\
            yelp                                & \itshape 0.86             & 0.69                    & 0.69                         & 0.86                          & \bfseries 0.87                 & 0.69                    & 0.69            & 0.86                         & \bfseries 0.87               & 0.69           & 0.76             & \bfseries 0.85               & 0.83                        & 0.69           & 0.69             & \bfseries 0.87               & \bfseries 0.87               & \itshape 0.69 \\
            yoruba                              & \itshape 0.50             & 0.37                    & 0.37                         & \bfseries 0.45                & 0.42                           & \bfseries 0.44          & 0.41            & 0.42                         & 0.40                         & \bfseries 0.44 & 0.43             & \bfseries 0.44               & \bfseries 0.44              & 0.43           & 0.43             & \bfseries 0.44               & \bfseries 0.44               & \itshape 0.44 \\
            youtube                             & \itshape 0.93             & \bfseries 0.93          & \bfseries 0.93               & \bfseries 0.93                & 0.92                           & \bfseries 0.93          & 0.92            & 0.92                         & 0.92                         & \bfseries 0.93 & 0.91             & \bfseries 0.93               & \bfseries 0.93              & 0.91           & 0.91             & \bfseries 0.92               & \bfseries 0.92               & \itshape 0.88 \\
            \midrule
            wins/draws/losses                   & -                         & -                       & \bfseries 0/19/0             & \bfseries \underline{12/3/4}  & \bfseries \underline{13/2/4}   & -                       & \bfseries 4/9/6 & \bfseries \underline{13/3/3} & \bfseries \underline{12/1/6} & -              & \bfseries 11/4/4 & \bfseries \underline{11/5/3} & \bfseries \underline{9/7/3} & -              & \bfseries 0/19/0 & \bfseries \underline{13/2/4} & \bfseries \underline{16/1/2} & -             \\
            \bottomrule
        \end{tabular}
    }
\end{table*}

\twocolumn

\clearpage

\section{Additional experiments}\label{sec:additional-experiments}

\begin{figure}[h]
\centering
\includegraphics[width=0.75\linewidth]{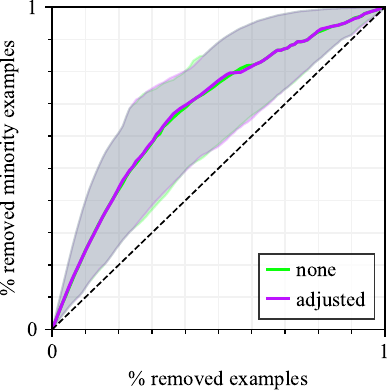}
\bottomcaption{Evolution of the number of removed examples when filtering instances from the least to the most trusted ones on the x-axis, with the number of removed examples from the minority classes (classes with priors less than $1/C$) on the y-axis. We plot the median ratio over all datasets alongside the 25\% and 75\% percentiles for detectors without calibration (none) and detectors with adjusted probabilities. Adjusting probabilities has no impact to help detectors differentiate minority examples from mislabeled ones, contrary to calibration (see Figure \ref{fig:minority_removed}).}
\label{fig:minority_removed_adjusted}
\end{figure}

\twocolumn

\end{document}